\documentclass{zapiski}

\usepackage[cp1251]{inputenc}
\usepackage[T1,T2A]{fontenc}
\usepackage[russian,english]{babel}
\usepackage{placeins}

\usepackage{amsmath}
\usepackage{amssymb}
\usepackage{bbm}

\usepackage{booktabs}
\usepackage{multirow}
\usepackage{makecell}

\usepackage{graphicx}
\usepackage{float}
\usepackage{capt-of}

\usepackage{listings}

\usepackage{url}
\usepackage{hyperref}

\newcommand{\sm}[1]{{\scriptsize$\pm$#1}}

\english  

\begin{document}

\renewcommand{\datename}{\fontencoding{T2A}\selectfont Поступило}
\originfo{529}{}{}{2023}
\setcounter{page}{177}
\date{\fontencoding{T2A}\selectfont 12 октября 2023 г.}

\title[Personal Facts Annotation]{An Annotation Scheme and Classifier for Personal Facts in Dialogue}

\author[K.~Zaitsev]{Konstantin Zaitsev}
\address[K.~Zaitsev]{HSE University, Moscow, Russia}
\email{kzaytsev@hse.ru}

\begin{abstract}
The advancement of Large Language Models (LLMs) has enabled their application in personalized dialogue systems. We present an extended annotation scheme for personal fact classification that addresses limitations in existing approaches, particularly PeaCoK. Our scheme introduces new categories (Demographics, Possessions) and attributes (Duration, Validity, Followup) that enable structured storage, quality filtering, and identification of facts suitable for dialogue continuation. We manually annotated 2,779 facts from Multi-Session Chat and trained a multi-head classifier based on transformer encoders. Combined with the Gemma-300M encoder, the classifier achieves $81.6 \pm 2.6$\% macro F1, outperforming all few-shot LLM baselines (best: GPT-5.4-mini, 72.92\%) by nearly 9 percentage points while requiring substantially fewer computational resources. Error analysis reveals persistent challenges in semantic boundary disambiguation, temporal aspect interpretation, and pragmatic reasoning for followup assessment. The dataset\footnotemark[1] and classifier\footnotemark[2] are publicly available.
\end{abstract}


\maketitle

\footnotetext[1]{Dataset: \href{https://huggingface.co/datasets/adugeen/personal-facts-msc}{HuggingFace dataset card}.}
\footnotetext[2]{Classifier: \href{https://huggingface.co/adugeen/personal-facts-classifier-embeddinggemma-300m}{HuggingFace model card}.}
\setcounter{footnote}{2}

\section{Introduction}

The advancement of Large Language Models (LLMs) has enabled their application in personalized dialogue systems~\cite{tseng-etal-2024-two}. Modern systems predominantly rely on LLMs that utilize prompting techniques~\cite{liu2025surveypersonalizedlargelanguage}. These systems address a diverse range of tasks, including everyday queries, recommendations, specialized domains such as programming or mathematics, as well as general conversational interaction that simulates human-like communication~\cite{10.1145/3771090}.

Through interaction with a dialogue system, a user provides personal facts related to themselves. Such information accumulates gradually, constructing a user profile, or persona. Persona may include the user's preferences, relationships with other persons, and past experiences. As a result of the active daily usage of dialogue systems, accumulated facts may be used to generate more personalized responses, thereby increasing engagement in the interaction.

The usage of personal facts in communication is directly related to the concepts of short-term and long-term memory. Several works explicitly distinguish between these two types of memory in dialogue systems~\cite{packer2024memgptllmsoperatingsystems, Zhong_Guo_Gao_Ye_Wang_2024, li-etal-2025-hello}. The key difference lies in the scope of information utilization. Short-term memory typically operates within the current dialogue context; the information it contains can be used to understand the ongoing communicative situation, and there is no need to store it in external memory banks. Long-term memory, on the other hand, relies on personal facts retrieved from external storage. Such information is utilized when the dialogue system needs to leverage personal knowledge after a certain period of time --- for instance, to establish a particular communicative context, assist in forming a user's profile, continue the dialogue on a previously discussed topic, or answer questions about the user and other persons related to them. While short-term memory does not necessarily require explicit extraction of user facts, long-term memory necessitates the implementation of a storage mechanism for personal information. This necessity is driven by several factors.

\begin{table}[t]
\centering
\small
\caption{Examples of annotated personal facts illustrating the proposed classification scheme. Abbreviations: Pref = Preferences, Goal = Goals and Plans, Exp = Experience, Rel = Relationships, Dem = Demographics, Pres = Present, Fut = Future, Dur = Duration, FU = Followup.}
\label{tab:examples_intro}
\resizebox{\textwidth}{!}{%
\begin{tabular}{p{3.5cm}llllll}
\toprule
\textbf{Personal Fact} & \textbf{Main Category} & \textbf{Time} & \textbf{Referent} & \textbf{Duration} & \textbf{Followup} & \textbf{Valid} \\
\midrule
I love Italian food. & Pref & Pres & Self & Long & -- & Yes \\
I'm going to Paris next week. & Goal & Fut & Self & Short & Yes & Yes \\
I worked at Google for five years. & Exp & Past & Self & Long & -- & Yes \\
My sister is a doctor. & Rel & Pres & Other & Long & -- & Yes \\
I have two cats and a dog. & Poss & Pres & Self & Long & -- & Yes \\
I want to learn Japanese someday. & Goal & Fut & Self & Long & Maybe & Yes \\
\midrule
The weather is nice today. & -- & -- & -- & -- & -- & No: No Fact \\
I'm 30 and I live in Boston. & -- & -- & -- & -- & -- & No: Multiple Facts \\
Speaker 1 does airplane work. & -- & -- & -- & -- & -- & No: Unattributable \\
\bottomrule
\end{tabular}%
}
\end{table}

First, the accumulation of facts leads to contextual noise~\cite{jin2024longcontextllmsmeetrag, chen2023benchmarkinglargelanguagemodels, fang-etal-2024-enhancing}. There are numerous approaches to utilizing information for question answering, with Retrieval Augmented Generation (RAG) being the most popular among them~\cite{lewis2021retrievalaugmentedgenerationknowledgeintensivenlp}. When accessing memory through either prompts or ranking models, the fact necessary for generating a response may be lost among other information, leading to decreased answer accuracy. Moreover, not all facts are applicable for maintaining long-term communication---some information may only be relevant at the current moment and never again. For instance, facts expressing user intentions that are expected to occur in the near future fall into this category, such as ``I'm going to a concert tomorrow.'' Such information is not essential for long-term retention, yet it remains important for a very short time span---the dialogue system may ask the user about how the concert went after some time has passed (the following day in the given example).

Second, there are limitations on the effective use of context and its size. Effective context refers to the ability of LLMs to locate relevant information within the context window. Notably, the context can be quite extensive. Numerous studies have examined the effectiveness of context utilization in modern LLMs~\cite{liu-etal-2024-lost, hsieh2024ruler, kuratov2024babilong}. In particular, Liu et al.~\cite{liu-etal-2024-lost} demonstrate that performance degrades significantly when relevant information is positioned in the middle of the input, while Hsieh et al.~\cite{hsieh2024ruler} show that most models claiming 32K+ context sizes fail to maintain satisfactory performance at those lengths.

The aforementioned factors necessitate addressing the problem of storing personal information in long-term memory. A review of related works revealed the absence of annotated data that systematically classifies facts. The exception is the dataset presented in PeaCoK~\cite{gao-etal-2023-peacok}. However, the annotation in that paper is designed for classifying relationships between facts rather than classifying the facts themselves. Thus, to bridge the gap we propose classifying and structuring personal facts developing an annotated dataset.

The contributions of this paper are as follows:
\begin{enumerate}
    \item \textbf{A developed annotation scheme for personal facts.} We propose classifying facts across seven dimensions: main category, time, referent, duration, validity, invalidity reason, and followup potential. Table~\ref{tab:examples_intro} presents annotated samples from the dataset.
    \item \textbf{A manually annotated dataset.} Results from preliminary annotation using language models revealed that they frequently make errors in certain categories. For this reason, manual annotation was conducted to achieve higher annotation quality.
    \item \textbf{Systematic ablation experiments.} We conduct encoder comparison experiments with the multi-head classification architecture, evaluating performance across 5 random seeds with proper train/validation/test splits.
    \item \textbf{A large-scale analysis of persona fact distributions.} We apply the trained classifier to PersonaChat (6,126 facts) and Multi-Session Chat (55,795 facts), providing a multi-dimensional characterization of personal facts in these widely-used datasets. The analysis estimates the prevalence and composition of potentially noisy MSC facts, with model predictions dominated by recoverable Multiple Facts cases, and identifies underrepresented categories affecting downstream modeling.
\end{enumerate}

\section{Related Work}
\label{sec:related_work}

The personalization of dialogue systems has become a pivotal area of research, with several studies leveraging user information---through persona concatenation~\cite{zhang2018personalizingdialogueagentsi}, persona-specific embeddings~\cite{huang-etal-2024-selective,tang2023enhancingpersonalizeddialoguegeneration}, RAG-based~\cite{huang-etal-2023-learning,liu-etal-2023-recap} and agentic methods~\cite{li-etal-2025-hello}---to enhance engagement and satisfaction~\cite{tseng-etal-2024-two,liu2025surveypersonalizedlargelanguage}. In this section, we review persona-based dialogue datasets, structured persona annotation approaches, multi-label classification methods, and the use of LLMs for data annotation.

\subsection{Dialogue Datasets}

Several datasets support persona-based dialogue research. PersonaChat~\cite{zhang2018personalizingdialogueagentsi} introduced persona-based conversations through a two-stage crowdsourcing process: participants first generated personal fact profiles, then paraphrased them to increase linguistic diversity (e.g., ``I like basketball'' $\rightarrow$ ``I am a fan of Michael Jordan''), and finally engaged in role-playing dialogues based on assigned personas. Multi-Session Chat (MSC)~\cite{xu-etal-2022-beyond} extends PersonaChat by simulating realistic long-term conversations spanning 3--5 sessions with temporal gaps, averaging 53--66 utterances per dialogue---substantially exceeding the length of comparable dialogue datasets available at the time of
publication. However, our analysis revealed that MSC contains a substantial proportion of invalid facts: those with unresolved object-level references (e.g., ``I've never read that book''---no antecedent for \emph{that book}), unattributable utterances such as third-person fragments referring to anonymized speakers (e.g., ``Speaker 1 does airplane work'') or assertions about non-personal entities (e.g., ``Chris Farley was only 33''), and overly general information. Such invalidity can degrade both persona modeling and fact extraction performance, highlighting the need for systematic quality assessment.

\subsection{Structured Persona Annotation}

The most closely related work on persona fact annotation is PeaCoK (Persona Commonsense Knowledge)~\cite{gao-etal-2023-peacok}, a large-scale knowledge graph containing approximately 100K human-validated persona facts. PeaCoK schematizes five primary dimensions of persona knowledge: \textit{characteristics} (intrinsic traits), \textit{routines and habits} (regular behaviors), \textit{goals and plans} (future intentions), \textit{experiences} (past events), and \textit{relationships} (social connections), along with supplementary axes including \textit{temporality}, \textit{regularity}, \textit{interactivity}, \textit{distinguishability}, and \textit{state}. Crucially, PeaCoK's annotation operates on \emph{relationships between fact pairs} rather than on individual facts---for instance, annotating the semantic connection between ``I sing'' and ``I want to be a famous singer.'' While this relationship-centric approach is well-suited for knowledge graph implementations, it presents practical challenges for modern LLM-based dialogue systems: knowledge graphs require specialized format and specific mechanisms for LLM context integration~\cite{Pan_2024,fatemi2024talk,deng2024graphvis}.

More critically, PeaCoK's taxonomy does not comprehensively cover all categories of personal facts arising in natural dialogue; facts expressing static properties such as possession (``I have a car'') or demographic attributes (``I am 25 years old'') do not naturally align with relationship-based classification. Our annotation scheme addresses these limitations by operating directly on individual facts, expanding the primary category taxonomy to include \textit{Possessions} and \textit{Demographics}, and introducing three practical dimensions absent from PeaCoK: \textit{Validity} (with detailed subcategories for data quality assessment), \textit{Duration} (short-term vs.\ long-term for memory management), and \textit{Followup} (identifying facts suitable for dialogue continuation). A more detailed comparison is presented in Appendix~\ref{sec:peacok_comparison}.

\subsection{Multi-Label Text Classification}

The task of assigning multiple labels to a single text instance has been extensively studied, with traditional approaches such as Binary Relevance~\cite{Tsoumakas2007Multi-Label} and Classifier Chains~\cite{10.5555/3121646.3121664}. Pre-trained transformer models have since become dominant: Devlin et al.~\cite{devlin-etal-2019-bert} demonstrated that fine-tuning BERT with task-specific classification heads achieves strong performance across diverse tasks. Recent work has explored sequence generation for label correlations~\cite{yang-etal-2018-sgm}, extreme multi-label methods~\cite{you2019attentionxmllabeltreebasedattentionaware}, and hierarchy-aware architectures for parent-child label dependencies~\cite{zhou-etal-2020-hierarchy}. While these methods have been applied to domains such as legal document classification~\cite{chalkidis-etal-2019-large} and biomedical text categorization~\cite{rios-kavuluru-2018-shot}, their application to persona fact classification remains underexplored. Our work extends this line of research by implementing specialized multi-head architectures tailored to the hierarchical structure and label dependencies present in personal fact annotation.

\subsection{LLM Annotation}

Recent advances in LLMs have opened new possibilities for scalable data annotation. Gilardi et al.~\cite{doi:10.1073/pnas.2305016120} demonstrate that ChatGPT can achieve annotation quality comparable to crowdworkers on several tasks, while He et al.~\cite{he-etal-2024-annollm} introduce AnnoLLM with consistency mechanisms for improved reliability. However, LLM-based annotation has known limitations: Zhu et al.~\cite{zhu2023chatgptreproducehumangeneratedlabels} identify systematic biases in subjective tasks, and Ding et al.~\cite{ding-etal-2023-gpt} find that GPT models struggle with fine-grained distinctions. In our work, we use LLMs (GPT-5-mini~\cite{singh2025openaigpt5card}, Gemma-3~\cite{gemmateam2025gemma3technicalreport}, Qwen-3~\cite{yang2025qwen3technicalreport}) for preliminary annotation, but rely on human labels to address LLM limitations. Our analysis shows that LLMs agree reasonably with humans on some categories (e.g., Main Category, Time), but struggle with fine-grained validity assessment, validating the need for human oversight.

\section{Dataset Creation}

The limitations identified in Section~\ref{sec:related_work}---namely, the prevalence of invalid facts in MSC and the insufficient coverage of PeaCoK's annotation categories---motivate the creation of a new annotated dataset. The dataset is designed to address these shortcomings by introducing a comprehensive annotation schema that captures multiple complementary dimensions of personal facts. In this section, we describe the proposed annotation schema in detail, outline the annotation guidelines used for both human annotators and LLMs, and examine the annotation results, highlighting the weaknesses of LLM-based annotation and inter-annotator inconsistencies.

\subsection{Annotation Scheme}

The annotation schema is guided by three practical requirements for long-term dialogue systems: (1)~structuring facts for efficient retrieval, (2)~identifying invalid or noisy facts that may degrade persona modeling, and (3)~determining which facts can be leveraged in future turns to enhance personalization. To address these requirements, we categorize personal facts along seven dimensions: Main Category, Time, Referent, Duration, Validity, Invalidity Reason, and Followup. Each dimension captures a distinct aspect of a personal fact, and together they support both memory management and dialogue planning.

\textbf{Main Category.} The main category describes the thematic content of personal facts, encompassing user preferences, past experiences, relationships, habits, routines, and demographic characteristics. Based on preliminary annotation, we identified the following classes:

\begin{enumerate}
    \item \textbf{Preferences.} Likes or dislikes toward activities, objects, or events. These signals are useful when the dialogue system offers recommendations.

    \item \textbf{Characteristics.} Internal qualities that constitute the user's persona: skills, distinctive traits, personality, knowledge, and social roles.

    \item \textbf{Routine Activities.} Actions performed on a regular basis, primarily habits and daily routines. While similar to preferences, this category specifically captures recurring behaviors rather than attitudes.

    \item \textbf{Experience.} Events that have occurred to the user in the past.

    \item \textbf{Goals and Plans.} Desires, objectives, and anticipated future events.

    \item \textbf{Relationships.} Facts characterizing the nature of the user's interpersonal connections, capturing sentiments, dynamics, and relational states (e.g., ``I hate my mother'' or ``My boss and I get along well'').

    \item \textbf{Demographics.} Demographic characteristics including age, gender, marital status, presence of children, and name.

    \item \textbf{Possessions.} Ownership or keeping of possessions, including personal items, real estate, vehicles, and pets.
\end{enumerate}

Classes 2--6 (Characteristics, Routine Activities, Experience, Goals and Plans, and Relationships) were adopted from PeaCoK, while Preferences was separated from PeaCoK's broader Characteristics dimension as a distinct category. Classes 7 and 8 (Demographics and Possessions) extend the existing annotation scheme. These additions address facts that proved difficult to categorize during preliminary annotation---particularly those where users state their name, mention children, or describe property ownership.

\textbf{Time.} This category characterizes personal facts from a temporal perspective, serving to determine fact relevance. Past-tense facts primarily constitute the user profile and are less likely to drive dialogue continuation. Present-tense facts also build the persona but can additionally introduce conversation topics; for instance, ``I run every day'' might prompt ``How was your run today?'' Future-tense facts reveal user goals and plans, and can similarly generate topics or elicit new information. Accordingly, three classes are defined: past, present, and future.

\textbf{Referent.} Utterances frequently contain facts about people other than the speaker. Distinguishing the referent is essential for long-term memory, which must store information attributed to specific individuals. Two classes are defined: Self and Other.

The Other class could be extended in future work to identify specific referents. For example, ``My mother likes watching movies'' would be attributed to ``My mother'' as a distinct entity. While this granularity falls outside the current scope, it represents a promising direction for annotation refinement.

\textbf{Duration.} Facts vary in their temporal relevance. Short-term facts apply only to the current or a limited time period; storing them in long-term memory creates unnecessary noise once they become outdated. Distinguishing between short-term and long-term facts therefore supports efficient memory management.

\textbf{Validity and Invalidity Reason.} Not all utterances labeled as personal facts in existing datasets constitute valid, extractable information. Some entries are invalid due to annotation artifacts or inherent ambiguity, though a subset may be recoverable through targeted refinement. We define the following invalidity categories:

\begin{enumerate}
    \item \textbf{No Fact.} Utterances lacking personal facts, typically containing general information unrelated to the user (e.g., ``Hippos can be aggressive'').

    \item \textbf{Opinion.} Expressions of opinion that do not directly contribute to the user profile, though implicit facts may sometimes be derivable.

    \item \textbf{Context Insufficient.} Facts whose meaning depends on an antecedent that cannot be recovered from the available context---typically unresolved object-level deixis (e.g., ``I've never read that book''---the referent of \emph{that book} is unrecoverable, even though the speaker is identified).

    \item \textbf{Unattributable.} Utterances that cannot be attributed to the speaker or any personally known individual. This subsumes three patterns observed in MSC: third-person fragments where the subject is an anonymized speaker placeholder rather than a known person (e.g., ``Speaker 1 does airplane work''), subject-less or sentence-fragment utterances (e.g., ``Drinking diet coke''), and assertions about public or non-personal entities (e.g., ``Chris Farley was only 33'').

    \item \textbf{Multiple Facts.} Utterances containing several facts belonging to different categories. Atomicity minimizes the risk of extracting irrelevant information; thus, ``I am 25 and I work at a factory'' should be split into ``I am 25'' (demographics) and ``I work at a factory'' (routine activities).
\end{enumerate}

Invalidity reasons are annotated only when a fact is marked as invalid. This subdivision enables targeted mitigation strategies: some facts are definitively unusable (No Fact, Opinion), while others can be refined through rephrasing or recovered with additional context (Context Insufficient, Unattributable, Multiple Facts).

\textbf{Followup.} This category identifies facts suitable for initiating future interactions, particularly relevant for future-tense statements. Consider two examples: ``I'm going to a concert tomorrow'' clearly supports a follow-up question (``How was your concert?''), whereas ``I plan to enroll in a medical program'' presents an indefinite timeline---weeks, years, or possibly never---making immediate follow-up inappropriate. Accordingly, we define two positive labels: \textit{Yes}, indicating high certainty that the fact can be referenced in a future interaction, and \textit{Maybe}, indicating that follow-up is plausible but contingent on uncertain conditions such as timing or user intent.

Combined with the Time category, Followup enables information elicitation: after sufficient time has passed, the dialogue system might inquire about enrollment progress, potentially yielding additional user information. This mechanism enhances both engagement and perceived personalization.

\medskip
\noindent In summary, the proposed scheme addresses three objectives:
\begin{enumerate}
    \item Structuring facts thematically and temporally.
    \item Detecting invalid facts and categorizing their deficiencies.
    \item Identifying facts with potential for dialogue continuation or information extraction.
\end{enumerate}

\subsection{Annotation Process}

\paragraph{Fact Sampling.} To ensure topical diversity and avoid over-representation of frequent fact types, we apply a cluster-based sampling strategy. Starting from the Multi-Session Chat corpus, we extract all personal facts associated with dialogue responses and deduplicate them by exact text match. Each unique fact is then embedded using BGE-M3~\cite{chen-etal-2024-m3} (dense vectors, L2-normalized) and clustered into 1,000 groups via K-Means. Finally, we randomly sample up to 3 facts per cluster, yielding 2,779 candidate facts for annotation. Compared to uniform random sampling, this approach ensures broader coverage of rare but semantically distinct fact types (e.g., relationships, possession of unusual items), producing a more representative dataset.

\paragraph{Annotation Guidelines.} Based on the proposed annotation scheme, we developed a detailed instruction for annotation using LLMs (see Appendix~\ref{appendix:annotation_prompt}). The instruction defines the unit of annotation as a single personal fact about the speaker or their personally known individuals. It specifies all category definitions with concrete examples, establishes priority rules for resolving ambiguous cases (e.g., Experience takes precedence for completed past events; Characteristics supersedes Demographics when phrased as identity), and provides explicit guidance for handling negation, modality, and first-person plural pronouns. The instruction also includes a step-by-step algorithm: first checking for validity issues, then determining the referent, selecting applicable categories, and finally assigning temporal and durational attributes. The annotation prompt additionally elicited \textit{specificity} (Specific/General) and \textit{context sufficiency} (Yes/No) judgments. However, specificity was excluded from the final classification task due to poor inter-annotator delimitation during preliminary trials, while context sufficiency was subsumed into the Invalidity Reason dimension as the ``Context Insufficient'' subcategory. The final classification dataset uses a canonical single-label representation: only the priority-resolved \texttt{main\_category} is retained, dual-duration cases are treated as ambiguous and excluded from supervised training and evaluation, and the rare \texttt{followup=No} outputs are folded into \textit{Maybe}. The prompt employs the field names \texttt{broken} and \texttt{broken\_reason}, which correspond to Validity and Invalidity Reason, respectively, in the terminology adopted throughout this paper. To ensure consistent output, annotators produce structured JSON with predefined keys and enumerated values. Four representative examples accompany the instruction to illustrate expected annotations across different fact types.

As annotator models, we propose using open-weight models and one proprietary model. Among the most relevant open-weight models, Gemma-3-27B\footnote{\url{https://huggingface.co/google/gemma-3-27b-it}} and Qwen-3-VL-32B\footnote{\url{https://huggingface.co/Qwen/Qwen3-VL-32B-Instruct}} can be highlighted. GPT-5-mini\footnote{\url{https://developers.openai.com/api/docs/models/gpt-5-mini}} is used as the proprietary model due to its favorable ratio of token cost to quality in preliminary annotation. Subsequently, the obtained annotation formed the basis for manual annotation. The annotator was provided with instructions and preliminary annotated classes, which are presented in the appendix. Multi-Session Chat was selected as the source dataset for annotation. Invalid facts are frequently encountered in MSC, and in general, the facts are atomized. These factors served as the reason for this choice. Manual annotation was performed by the author using label-studio\footnote{\url{https://labelstud.io}}.

\subsection{Annotation Result}

\begin{table}[t]
\centering
\footnotesize
\setlength{\tabcolsep}{4pt}
\caption{Dataset statistics. Label distribution across annotation categories (2,779 examples total). Time, Referent, and Duration percentages do not sum to 100\% because invalid facts receive a ``None'' label for these dimensions.}
\label{tab:dataset_stats}
\begin{tabular}{llrc}
\toprule
\textbf{Category} & \textbf{Label} & \textbf{Count} & \textbf{\%} \\
\midrule
\multirow{9}{*}{Main}
& Preferences & 573 & 20.6 \\
& None & 423 & 15.2 \\
& Experience & 385 & 13.9 \\
& Routine Activities & 373 & 13.4 \\
& Goals and Plans & 336 & 12.1 \\
& Characteristics & 288 & 10.4 \\
& Demographics & 180 & 6.5 \\
& Possessions & 178 & 6.4 \\
& Relationships & 43 & 1.5 \\
\midrule
\multirow{3}{*}{Time}
& Present & 1,663 & 59.8 \\
& Past & 363 & 13.1 \\
& Future & 329 & 11.8 \\
\midrule
\multirow{2}{*}{Referent}
& Self & 2,166 & 77.9 \\
& Other & 189 & 6.8 \\
\midrule
\multirow{2}{*}{Duration}
& Long-term & 1,884 & 67.8 \\
& Short-term & 471 & 16.9 \\
\midrule
Validity & Valid & 2,356 & 84.8 \\
\midrule
\multirow{5}{*}{Inv. Reason}
& Opinion & 115 & 4.1 \\
& Multiple Facts & 94 & 3.4 \\
& No Fact & 92 & 3.3 \\
& Unattributable & 79 & 2.8 \\
& Context Insuff. & 44 & 1.6 \\
\midrule
\multirow{2}{*}{Followup}
& Yes & 203 & 7.3 \\
& Maybe & 148 & 5.3 \\
\bottomrule
\end{tabular}
\end{table}

\begin{table}[t]
\centering
\caption{Inter-annotator agreement between human annotations and LLM-based annotations (GPT-5-mini, Gemma-3-27B, Qwen-3-VL-32B). Fleiss' $\kappa$ is reported for overall multi-rater agreement, and Krippendorff's $\alpha$ for nominal data. Interpretation labels follow Landis and Koch~\cite{landis-koch-1977}.}
\label{tab:interannotator_agreement}
\resizebox{\textwidth}{!}{%
\begin{tabular}{lccccr}
\toprule
\textbf{Class} & \textbf{\% Agreement} & \textbf{Fleiss' $\kappa$} & \textbf{Interpretation} & \textbf{Krippendorff's $\alpha$} & \textbf{N} \\
\midrule
Main Category       & 78.6\% & 0.749 & Substantial     & 0.748 & 2,732 \\
Time                & 88.3\% & 0.786 & Substantial     & 0.791 & 2,732 \\
Referent            & 88.1\% & 0.630 & Substantial     & 0.619 & 2,732 \\
Duration            & 86.2\% & 0.670 & Substantial     & 0.650 & 2,732 \\
Validity            & 92.1\% & 0.549 & Moderate        & 0.540 & 2,732 \\
Invalidity Reason   & 90.1\% & 0.458 & Moderate        & 0.450 & 2,732 \\
Followup            & 94.1\% & 0.758 & Substantial     & 0.765 & 2,732 \\
\midrule
\textbf{Average}    & \textbf{88.2\%} & \textbf{0.657} & \textbf{Substantial} & \textbf{0.652} & \\
\bottomrule
\end{tabular}%
}
\end{table}

\begin{table}[t]
\centering
\caption{Pairwise human-human agreement on a 315-example validation subset. Cohen's $\kappa$ is reported for pairwise agreement, and Krippendorff's $\alpha$ for nominal data. Interpretation labels follow Landis and Koch~\cite{landis-koch-1977}.}
\label{tab:human_human_agreement}
\resizebox{\textwidth}{!}{%
\begin{tabular}{lccccr}
\toprule
\textbf{Class} & \textbf{\% Agreement} & \textbf{Cohen's $\kappa$} & \textbf{Interpretation} & \textbf{Krippendorff's $\alpha$} & \textbf{N} \\
\midrule
Main Category       & 78.1\% & 0.748 & Substantial & 0.748 & 315 \\
Time                & 86.0\% & 0.754 & Substantial & 0.754 & 315 \\
Referent            & 88.9\% & 0.677 & Substantial & 0.677 & 315 \\
Duration            & 81.9\% & 0.617 & Substantial & 0.617 & 315 \\
Validity            & 90.8\% & 0.670 & Substantial & 0.669 & 315 \\
Invalidity Reason   & 87.6\% & 0.587 & Moderate    & 0.586 & 315 \\
Followup            & 91.4\% & 0.530 & Moderate    & 0.529 & 315 \\
\midrule
\textbf{Average}    & \textbf{86.4\%} & \textbf{0.655} & \textbf{Substantial} & \textbf{0.654} & \\
\bottomrule
\end{tabular}%
}
\end{table}

The final annotated dataset comprises 2,779 examples (Table~\ref{tab:dataset_stats}). Several patterns emerge from the annotation:

\textbf{Main Category.} The distribution is relatively balanced. Preferences (20.6\%) and Experience (13.9\%) are most frequent, while Relationships (1.5\%) is least represented---consistent with its lower occurrence in natural dialogue.

\textbf{Time.} Present tense dominates (59.8\%), reflecting typical conversational patterns where speakers discuss current states and ongoing activities.

\textbf{Referent.} Facts predominantly concern the speaker (77.9\%), with only 6.8\% referring to other people. This validates the self-focused nature of persona construction in dialogue.

\textbf{Duration.} Long-term facts (67.8\%) significantly outnumber short-term ones (16.9\%), indicating that most personal information carries enduring relevance for user profiling.

\textbf{Validity.} Valid facts constitute 84.8\% of the dataset, while 15.2\% are marked invalid. Among invalid instances, Opinion (4.1\%) and Multiple Facts (3.4\%) predominate, followed by No Fact (3.3\%).

\textbf{Followup.} Only 12.6\% of facts support follow-up conversations (7.3\% definite, 5.3\% tentative), suggesting that most facts serve profile construction rather than actionable dialogue continuation.

\subsection{Inter-Annotator Agreement}

Table~\ref{tab:interannotator_agreement} presents agreement across four annotators: one human annotator and three LLMs, computed over 2,732 jointly annotated examples.\footnote{The 47 examples (1.7\%) excluded from IAA computation are those for which at least one LLM did not produce an aligned annotation (e.g., output that failed to match the canonical (context, text) key of the human annotation); they are retained in the human-annotated dataset of 2,779 facts.} The interpretation labels use the Landis and Koch scale~\cite{landis-koch-1977}. The results show substantial overall agreement (Fleiss' $\kappa$ = 0.657, Krippendorff's $\alpha$ = 0.652), with five of seven categories reaching substantial reliability. Followup achieves the highest agreement (94.1\%, $\kappa$ = 0.758), while Referent (88.1\%, $\kappa$ = 0.630) and Time (88.3\%, $\kappa$ = 0.786) also show strong consensus.

To further assess the reliability of the human annotation, we additionally collected independent second-human labels for a 315-example validation subset. Table~\ref{tab:human_human_agreement} reports pairwise agreement between the original author annotation and the independent annotator. The average human-human agreement is 86.4\%, with Cohen's $\kappa$ = 0.655 and Krippendorff's $\alpha$ = 0.654 across the seven dimensions, closely matching the aggregate human--LLM agreement while providing a direct check that the annotation scheme is reproducible across human annotators. Agreement is strongest for Time, Referent, and Validity, while Followup and Invalidity Reason remain comparatively more subjective.

Two categories exhibit lower reliability. Invalidity Reason achieves only moderate agreement (Fleiss' $\kappa$ = 0.458, Krippendorff's $\alpha$ = 0.450), the lowest across all dimensions. While annotators can identify obviously valid facts (Validity surface agreement reaches 92.1\%), explaining \textit{why} a fact is invalid---distinguishing between ``No Fact,'' ``Opinion,'' ``Context Insufficient,'' ``Unattributable,'' and ``Multiple Facts''---is more difficult. This finding aligns with prior observations that fine-grained classification requiring nuanced reasoning remains difficult for both LLMs and human annotators~\cite{ding-etal-2023-gpt}. Main Category, despite reaching substantial agreement ($\kappa$ = 0.749), has the lowest surface agreement (78.6\%), reflecting genuine difficulty in distinguishing semantically overlapping categories. For instance, differentiating ``I enjoy playing guitar'' (Preferences vs.\ Routine Activities vs.\ Characteristics) requires subtle semantic judgment.

Together, these results validate two key methodological decisions. First, manual annotation with human oversight remains essential, particularly for validity assessment where agreement is lowest. Second, the moderate agreement on Invalidity Reason justifies our decision to train specialized classifiers rather than relying solely on LLM few-shot predictions, as even combined human-LLM annotation struggles with the precision required for nuanced data quality assessment.

\section{Methods}

Since each personal fact requires annotation across multiple categories simultaneously, we frame this as a multi-output classification problem. We propose a multi-head classification architecture built on a transformer encoder backbone~\cite{vaswani2023attentionneed}.

\subsection{Multi-Head Classifier}

The Multi-Head Classifier employs separate classification heads for each target category. Given input text $x$, a pre-trained transformer encoder produces a contextualized representation:
\begin{equation}
\mathbf{h} = \text{Encoder}(x) \in \mathbb{R}^d
\end{equation}
where $\mathbf{h}$ corresponds to the \texttt{[CLS]} token representation and $d$ is the hidden dimension.

For each category $c \in \{1, \ldots, C\}$, an independent classification head computes:
\begin{align}
\mathbf{z}_c &= \text{Dropout}(\mathbf{h}) \\
\tilde{\mathbf{z}}_c &= \tanh(\mathbf{W}_c^{(1)} \mathbf{z}_c + \mathbf{b}_c^{(1)}) \\
\boldsymbol{\ell}_c &= \mathbf{W}_c^{(2)} \text{Dropout}(\tilde{\mathbf{z}}_c) + \mathbf{b}_c^{(2)}
\end{align}
where $\boldsymbol{\ell}_c \in \mathbb{R}^{n_c}$ denotes the logits and $n_c$ is the number of labels for category $c$.

The training objective minimizes the weighted sum of cross-entropy losses:
\begin{equation}
\mathcal{L} = \frac{1}{|\mathcal{V}|} \sum_{c=1}^{C} w_c \cdot \mathbbm{1}[y_c \geq 0] \cdot \text{CE}(\boldsymbol{\ell}_c, y_c)
\end{equation}
where $y_c$ is the ground-truth label, $w_c$ is the class weight, and $\mathcal{V}$ denotes the set of valid (non-masked) categories.

This architecture offers three advantages: (1)~each head specializes in its respective category, (2)~it naturally accommodates varying label counts per category, and (3)~it enables independent confidence estimation for each prediction.

\subsection{Experimental Setup}

We evaluate the Multi-Head Classifier with several encoder models of comparable size: RoBERTa-large\footnote{\url{https://huggingface.co/FacebookAI/roberta-large}}~\cite{liu2019robertarobustlyoptimizedbert}, BGE-large-en-v1.5\footnote{\url{https://huggingface.co/BAAI/bge-large-en-v1.5}}~\cite{bge_embedding}, Gemma Embedding (300M)\footnote{\url{https://huggingface.co/google/embeddinggemma-300m}}~\cite{vera2025embeddinggemmapowerfullightweighttext}, and Multilingual-E5-base\footnote{\url{https://huggingface.co/intfloat/multilingual-e5-base}}~\cite{wang2024multilingual}. These encoders were selected based on their strong performance on multi-label tasks in the MTEB benchmark\footnote{\url{https://huggingface.co/spaces/mteb/leaderboard}}~\cite{enevoldsen2025mmtebmassivemultilingualtext} among models with similar parameter counts.

\paragraph{Training Details.} We use a 70/10/20 train/val/test split with stratification by main category. All models are optimized using AdamW with a learning rate of $2 \times 10^{-5}$ and batch size of 64. Training runs for 10 epochs on a single NVIDIA RTX 5090 GPU, with early stopping based on validation macro F1; the held-out test set is used only for final evaluation. To assess result stability, we repeat all experiments across 5 random seeds (42, 123, 456, 789, 1024), each producing a different stratified split and training initialization. We report mean and standard deviation of macro-averaged F1 scores.

\paragraph{Traditional Baseline Features.} The traditional ML baselines (SVM, Logistic Regression, Gradient Boosting) operate on TF-IDF features extracted with scikit-learn's \texttt{TfidfVectorizer}: 1--2 word $n$-grams, $\mathrm{min\_df}=2$, $\mathrm{max\_df}=0.95$, $\mathrm{max\_features}=10{,}000$, sublinear term-frequency scaling and Unicode accent stripping. SVM and Logistic Regression additionally use class-balanced sample weights to mitigate label skew; all baselines share the same train/val/test splits as the encoder models.

\paragraph{Deployment Constraint and Choice of Baselines.} We deliberately compare the fine-tuned small encoders against few-shot LLMs rather than fine-tuned LLMs. Fine-tuning a 27B+ LLM on 2{,}779 examples would likely improve its absolute numbers, but this is not the operational regime: a production dialogue system has to classify every incoming utterance in tens of milliseconds, which excludes 27B+ models regardless of fine-tuning cost. Self-consistency aggregation ($k\!\geq\!5$ samples) similarly multiplies inference cost and is impractical for per-turn fact classification. The reported baselines reflect this deployment constraint, and the headline comparison is intentionally a 300M fine-tuned encoder versus few-shot LLMs of $10$--$100\times$ the parameter count. In addition to the proprietary GPT-5-mini and GPT-5.4-mini baselines, we evaluate three open-weight models via OpenRouter---DeepSeek-V3.2, Qwen3-235B-A22B-2507, and Llama-3.3-70B-Instruct---to provide a broader comparison across model families.

\section{Results}

\begin{table}[t]
\centering
\caption{Performance comparison across encoders and baselines. Values are macro-averaged F1 scores (\%); encoder and traditional ML results are reported as mean\sm{std} over 5 random seeds, while few-shot LLM results are from a single evaluation on the held-out test set. Best result per column in \textbf{bold}, second best \underline{underlined}. MC = Main Category, Tm = Time, Ref = Referent, Dur = Duration, Val = Validity, IR = Invalidity Reason, FU = Followup, Ovr = Overall. Overall macro F1 is computed by pooling predictions across all seven categories into a single evaluation and computing macro-averaged F1 over all label types jointly, rather than averaging per-category F1 scores.}
\label{tab:model_comparison}
\resizebox{\textwidth}{!}{%
\begin{tabular}{lcccccccc}
\toprule
\textbf{Model} & \textbf{MC} & \textbf{Tm} & \textbf{Ref} & \textbf{Dur} & \textbf{Val} & \textbf{IR} & \textbf{FU} & \textbf{Ovr $\uparrow$}  \\
\midrule
\multicolumn{9}{l}{\textit{Few-Shot LLMs}} \\
GPT-5.4-mini & \underline{76.50} & 80.75 & 77.75 & 76.13 & 78.96 & \underline{53.81} & 72.13 & 72.92 \\
GPT-5-mini & 74.38 & 82.09 & 79.29 & 78.46 & 80.44 & 49.37 & \underline{79.00} & 72.20 \\
Llama-3.3-70B-Instruct & 71.96 & 82.11 & 82.48 & 76.20 & 54.24 & 49.46 & 71.44 & 70.41 \\
Qwen3-235B-A22B-2507 & 65.59 & 82.33 & 79.43 & 71.51 & 53.28 & 47.68 & 74.34 & 70.04 \\
DeepSeek-V3.2 & 70.36 & 79.59 & 73.47 & 74.46 & 74.12 & 44.50 & 70.67 & 68.69 \\
\midrule
\multicolumn{9}{l}{\textit{Multi-Head encoder comparison}} \\
Gemma-300M      & \textbf{79.4}\sm{2.5} & 85.7\sm{1.1} & \underline{84.7}\sm{1.0} & \underline{80.4}\sm{1.8} & \textbf{84.2}\sm{2.5} & \textbf{59.0}\sm{5.3} & 78.8\sm{2.5} & \textbf{81.6}\sm{2.6} \\
BGE-large       & 74.4\sm{2.0} & \underline{86.0}\sm{1.3} & \textbf{85.5}\sm{1.1} & \underline{80.4}\sm{2.0} & \textbf{84.2}\sm{2.0} & 53.5\sm{5.1} & 78.0\sm{2.1} & \underline{76.5}\sm{2.0} \\
RoBERTa-large   & 74.8\sm{2.8} & \textbf{86.2}\sm{1.5} & 83.2\sm{2.6} & \textbf{81.4}\sm{2.4} & 83.5\sm{2.0} & 48.3\sm{3.5} & \textbf{80.2}\sm{3.1} & \underline{76.5}\sm{2.5} \\
Multi-E5-base   & 67.1\sm{2.8} & 85.1\sm{1.6} & 82.0\sm{1.1} & 78.9\sm{2.2} & \underline{83.9}\sm{2.2} & 36.8\sm{3.0} & 77.1\sm{1.6} & 71.7\sm{1.6} \\
\midrule
\multicolumn{9}{l}{\textit{Traditional ML baselines}} \\
SVM             & 61.1\sm{3.0} & 72.9\sm{2.2} & 71.0\sm{4.4} & 63.3\sm{2.4} & 70.3\sm{1.0} & 40.0\sm{1.8} & 69.7\sm{2.4} & 61.1\sm{2.3} \\
Logistic Reg.   & 60.1\sm{0.4} & 71.1\sm{1.9} & 69.0\sm{2.4} & 64.3\sm{1.2} & 71.2\sm{1.6} & 40.3\sm{3.2} & 68.6\sm{1.9} & 60.2\sm{1.0} \\
Gradient Boost. & 57.2\sm{1.2} & 67.7\sm{1.7} & 64.8\sm{3.3} & 60.0\sm{3.6} & 69.8\sm{2.1} & 38.4\sm{5.2} & 64.7\sm{1.7} & 57.9\sm{1.7} \\
\bottomrule
\end{tabular}%
}
\end{table}

Table~\ref{tab:model_comparison} presents results from the encoder comparison and baseline evaluation. Encoder and traditional ML results are averaged over 5 random seeds, while few-shot LLM baselines are evaluated once on the held-out test set. The fine-tuned Gemma-300M encoder with the Multi-Head architecture achieves the highest overall performance ($81.6 \pm 2.6$\% macro F1), substantially outperforming traditional ML approaches. Gemma-300M demonstrates particularly strong performance on Main Category ($79.4 \pm 2.5$\%) and Invalidity Reason ($59.0 \pm 5.3$\%), indicating effective learning of the core taxonomic distinctions essential for persona fact classification. BGE-large-en-v1.5 achieves the highest Referent F1 ($85.5 \pm 1.1$\%), while RoBERTa-large leads on Time ($86.2 \pm 1.5$\%) and Followup ($80.2 \pm 3.1$\%), suggesting that different encoders capture complementary aspects of personal fact semantics.

Among the few-shot LLM baselines, GPT-5.4-mini and GPT-5-mini achieve the strongest overall performance (72.92\% and 72.20\% macro F1, respectively). GPT-5-mini attains the second-best Followup F1 (79.00\%) across all models, while GPT-5.4-mini leads the LLMs on Main Category (76.50\%) and Invalidity Reason (53.81\%). The open-weight models---Llama-3.3-70B-Instruct\footnote{\url{https://openrouter.ai/meta-llama/llama-3.3-70b-instruct}} (70.41\%), Qwen3-235B-A22B-2507\footnote{\url{https://openrouter.ai/qwen/qwen3-235b-a22b-2507}} (70.04\%), and DeepSeek-V3.2\footnote{\url{https://openrouter.ai/deepseek/deepseek-v3.2}} (68.69\%)---perform comparably to one another but trail the proprietary models by 2--4 percentage points overall. Notably, Llama-3.3-70B-Instruct achieves the highest LLM Referent F1 (82.48\%), approaching the fine-tuned encoders. The performance gap between all LLMs and the fine-tuned encoders is particularly pronounced for Invalidity Reason, where even the best LLM (GPT-5.4-mini, 53.81\%) falls short of Gemma-300M ($59.0 \pm 5.3$\%), underscoring the importance of supervised learning for nuanced data quality assessment.

\section{Discussion and Future Work}

This section examines the practical implications of our findings, analyzes systematic error patterns, and compares label distributions across existing persona datasets. Together, these analyses inform both deployment considerations and directions for future research.

\subsection{Implications for Practical Deployment}

Gemma-300M achieves $81.6 \pm 2.6$\% macro F1 with only 300M parameters, outperforming the best few-shot LLM baseline (GPT-5.4-mini, 72.92\%) by nearly 9 percentage points while requiring substantially fewer computational resources. For real-time classification in production dialogue systems, this efficiency advantage proves crucial: the model maintains inference speeds compatible with interactive applications while exceeding the accuracy of models orders of magnitude larger.

Our dataset analysis reveals substantial preprocessing opportunities in existing corpora, particularly the model-estimated $29.7 \pm 7.2$\% invalidity rate in MSC. This should not be read as a purely unrecoverable error rate: many predicted invalid facts are bundled Multiple Facts cases that can be decomposed into atomic statements, while facts with insufficient context can be augmented with dialogue history. Developing automated rephrasing models for such cases represents a promising direction. Additionally, the current annotation scheme could be extended with finer-grained attributes. Duration could be enriched with explicit temporal estimates (e.g., days, months, or years), enabling more precise memory eviction policies. Similarly, Referent could be extended to resolve specific entities (e.g., mapping ``my mother'' to a named individual), supporting structured persona graphs.

From a systems perspective, the proposed classification scheme maps naturally onto the memory architecture of LLM-based dialogue agents. Facts classified as long-term and valid can be serialized into structured storage (e.g., JSON) for persistent retrieval, while short-term facts and those marked with high followup potential can be maintained in a transient context window. Once short-term facts lose their temporal relevance, they can be evicted to reduce context usage---a strategy that finer-grained duration annotation would directly support. Integrating such schema-driven memory management into production dialogue systems constitutes a key direction for future work.

\subsection{Error Analysis}
\label{sec:error_analysis_section}

We inspected errors from our best model to identify systematic patterns. Per-label F1 scores across all categories are reported in Appendix~\ref{sec:per_label_f1} (Table~\ref{tab:per_label_f1}); here we highlight four primary issues.

\paragraph{Semantic Boundary Ambiguity.} The model frequently confuses categories with overlapping semantics. Relationships facts describing interpersonal dynamics indirectly (e.g., ``My parents do not understand how busy I am'') are misclassified as Characteristics when the model focuses on speaker state rather than relational content. Similarly, Preference-Characteristic boundaries prove difficult: facts like ``I love living on Earth'' and ``I like learning new languages'' blur the line between stable attributes and subjective evaluations---an ambiguity reflected in the lowest surface agreement (78.6\%) among all categories for Main Category.

\paragraph{Temporal and Aspectual Ambiguity.} Despite reasonable overall Time F1 ($85.7 \pm 1.1$\%), the None class within this dimension achieves only 71.14\% F1, suffering from tense interpretation errors, particularly with perfect aspect constructions. ``I have lived in Alaska'' and ``I have always had to share a room'' describe past events with ongoing relevance, making Past/Present distinctions unclear. The model also struggles with implicit temporal grounding: ``My grandparents are all deceased'' expresses past events through present state, while ``One of my ancestors is a civil war general'' requires inferring historical reference.

\paragraph{Possession-Attribution Bias.} Despite strong Referent performance ($84.7 \pm 1.0$\% F1), the model systematically misclassifies facts about named pets. ``My golden retriever is named Maggie and is nine years old'' was labeled Self (98.9\% confidence) rather than Other, suggesting the model treats possessive constructions as self-referential without recognizing that attributes of possessed entities refer to the Other. This bias has implications for persona graph construction, potentially causing incorrect fact attribution.

\paragraph{Pragmatic Reasoning Limitations.} Duration and Followup categories require pragmatic assessment that the model handles inconsistently. Long-term aspirations like ``I want to open a jiu-jitsu studio'' were classified as Short-term, while the model underestimated followup potential for concrete goals. Invalidity Reason achieves only $59.0 \pm 5.3$\% F1, with the model struggling to distinguish personal facts from general statements (e.g., ``I worry the summer heat may be too much''). This lowest-performing category may require human oversight or specialized models rather than joint classification.

\subsection{Dataset Distribution Analysis}

PersonaChat and Multi-Session Chat are among the most widely used resources for persona-grounded dialogue, yet the composition of their personal facts has not been systematically characterized beyond surface-level statistics. We apply the five trained seed models to PersonaChat (6,126 facts) and the full MSC corpus (55,795 facts) and report mean and standard deviation of predicted label distributions (Table~\ref{tab:dataset_distributions}). While PersonaChat shows stable predictions across seeds (most categories with $\sigma < 1$\%), MSC exhibits substantially higher variance ($\sigma$ up to 7.4\%), indicating that model predictions on multi-session dialogue generalize less reliably than on persona-focused exchanges. Given this high seed-to-seed variance, MSC distribution estimates should be interpreted as broad trends rather than precise point estimates.

\paragraph{Category Balance.} PersonaChat shows a relatively balanced Main Category distribution, with Preferences ($23.1 \pm 0.8$\%) and Characteristics ($21.7 \pm 1.1$\%) dominating. MSC exhibits a strikingly different profile: None becomes the largest single category ($30.4 \pm 6.4$\%), followed by Preferences ($23.7 \pm 3.3$\%), suggesting that a substantial fraction of MSC utterances do not convey classifiable personal facts. Both datasets severely underrepresent Relationships ($1.8 \pm 0.6$\% and $1.0 \pm 0.3$\%), limiting their utility for modeling interpersonal dynamics.

\paragraph{Temporal and Referential Skew.} Both datasets exhibit class imbalance, though the degree differs markedly. PersonaChat is dominated by Present tense ($81.3 \pm 1.2$\%) and Self-referent facts ($89.6 \pm 0.8$\%), consistent with its self-introductory format. MSC shows lower proportions for both (Present: $58.5 \pm 6.7$\%, Self: $66.3 \pm 6.7$\%), with the remainder largely attributed to the None category (${\sim}30$\% across Time, Referent, and Duration), reflecting the model's difficulty in assigning clear temporal or referential attributes to multi-session utterances.

\paragraph{Data Quality.} MSC is classified as containing substantially more invalid facts ($29.7 \pm 7.2$\%) than PersonaChat ($5.3 \pm 0.8$\%), though the wide confidence interval for MSC (individual seeds range from 21\% to 40\%) warrants cautious interpretation. Among Invalidity Reason predictions, Multiple Facts dominates MSC ($27.9 \pm 6.8$\%), suggesting that the main preprocessing opportunity is decomposing bundled facts from multi-session summaries. These invalidity rates are classifier predictions rather than gold annotations and should be read as model-estimated lower bounds rather than absolute measurements of MSC quality. These results underscore the need for careful preprocessing and quality filtering when using MSC for persona modeling.

\paragraph{Training-Data Leakage Caveat.} Because our 2{,}779 annotated facts were sampled from MSC, predictions on the full-MSC distribution are partly evaluated on examples the classifier has seen at training time. We quantified this exposure by matching annotated texts against the 55{,}795-fact MSC prediction set: 183 examples ($\approx\!0.33$\% of the full corpus) overlap with the training set. Re-aggregating Table~\ref{tab:dataset_distributions} with these 183 examples held out shifts every cell by less than $0.11$ percentage points (maximum shift: $0.109$~pp at \textit{Time / Present}). The reported MSC distributions should therefore be read as model-estimated trends rather than ground-truth measurements, but the leakage effect on the headline statistics is well within the seed-to-seed variance ($\sigma$ up to $7.4$\%) already noted above.

\paragraph{Followup Scarcity.} Both datasets show extremely low proportions of facts suitable for dialogue continuation (Yes: ${\sim}2.2$\%, Maybe: $3.0$--$3.4$\%). This severe imbalance presents challenges for training followup detection and may explain our model's lower performance on this category. Future dataset construction should specifically target facts with clear followup potential.

\section{Conclusion}

We presented an extended annotation scheme for personal fact classification that addresses limitations in existing approaches, particularly PeaCoK. Our scheme introduces new categories (Demographics, Possessions) and attributes (Duration, Validity, Followup) that enable structured storage, quality filtering, and identification of facts suitable for dialogue continuation. We manually annotated 2,779 facts from Multi-Session Chat and demonstrated that large language models struggle with this task, particularly for validity assessment---motivating the need for specialized classifiers.

We proposed a multi-head classification architecture and conducted encoder comparison experiments across 5 random seeds with train/val/test splits. Combined with Gemma-300M, the Multi-Head Classifier achieves $81.6 \pm 2.6$\% macro F1, outperforming the best few-shot LLM baseline by nearly 9 percentage points while requiring substantially fewer computational resources. Error analysis reveals persistent challenges in semantic boundary disambiguation, temporal aspect interpretation, and pragmatic reasoning for followup assessment. Application of our classifier to PersonaChat and the full MSC dataset reveals significant distributional differences, with MSC containing a model-estimated $29.7 \pm 7.2$\% of facts classified as invalid compared to PersonaChat's $5.3 \pm 0.8$\%; in both corpora, fewer than 6\% of facts support natural dialogue continuation.

\appendix

\FloatBarrier

\section{Annotation Comparison with PeaCoK}
\label{sec:peacok_comparison}

\begin{table}[H]
\centering
\caption{Comparison between PeaCoK's relationship-based annotation and our fact-level annotation scheme.}
\label{tab:peacock_comparison}
\scriptsize
\setlength{\tabcolsep}{2pt}
\begin{tabular}{@{}p{2.25cm}p{4.0cm}p{4.25cm}@{}}
\toprule
\textbf{Aspect} & \textbf{PeaCoK} & \textbf{Our Scheme} \\
\midrule
\textbf{Annotation Unit} & Fact pairs (relationships) & Individual facts \\
\midrule
\textbf{Primary Categories} &
5 relationship types:
\newline \textbullet{} Characteristics
\newline \textbullet{} Routines and Habits
\newline \textbullet{} Goals and Plans
\newline \textbullet{} Experiences
\newline \textbullet{} Relationships
&
8 fact types (adds 3 new):
\newline \textbullet{} Characteristics
\newline \textbullet{} Routine Activities
\newline \textbullet{} Goals and Plans
\newline \textbullet{} Experiences
\newline \textbullet{} Relationships
\newline \textbullet{} \textbf{Preferences} (new, from Characteristics)
\newline \textbullet{} \textbf{Possessions} (new)
\newline \textbullet{} \textbf{Demographics} (new) \\
\midrule
\textbf{Additional Dimensions} &
5 axes:
\newline \textbullet{} Temporality
\newline \textbullet{} Regularity
\newline \textbullet{} Interactivity
\newline \textbullet{} Distinguishability
\newline \textbullet{} State
&
6 dimensions:
\newline \textbullet{} Time (explicit temporal class)
\newline \textbullet{} Referent (Self/Other attribution)
\newline \textbullet{} \textbf{Validity} (data quality)
\newline \textbullet{} \textbf{Invalidity Reason} (subcategories)
\newline \textbullet{} \textbf{Duration} (memory management)
\newline \textbullet{} \textbf{Followup} (dialogue utility) \\
\midrule
\textbf{Example 1} &
Head: ``I am a singer''
\newline Relation: Characteristic
\newline Tail: ``good at singing''
\newline (requires 2 separate facts) &
Fact: ``I am a singer''
\newline Main Category: Characteristics
\newline (single self-sufficient fact) \\
\midrule
\textbf{Example 2} &
Head: ``I sing''
\newline Relation: Goal/Plan
\newline Tail: ``I want to be famous''
\newline (relationship between two facts) &
Fact: ``I want to be a famous singer''
\newline Main Category: Goals and Plans
\newline Followup: Yes
\newline (atomic fact with dialogue utility) \\
\midrule
\textbf{Example 3} &
Not naturally representable
\newline (possession not a relationship) &
Fact: ``I have a car''
\newline Main Category: Possessions
\newline (explicitly covered) \\
\midrule
\textbf{Example 4} &
Not naturally representable
\newline (age not relational) &
Fact: ``I am 25 years old''
\newline Main Category: Demographics
\newline Duration: Long-term
\newline (explicitly covered) \\
\bottomrule
\end{tabular}
\end{table}

Table~\ref{tab:peacock_comparison} contrasts PeaCoK's relationship-based annotation with our fact-level scheme along four axes (annotation unit, primary categories, additional dimensions, and worked examples), highlighting the categories and attributes that the proposed scheme adds.

\FloatBarrier

\section{Annotation Prompt for Language Models}
\label{appendix:annotation_prompt}

The following prompt was used to instruct large language models for preliminary annotation of personal facts. The prompt defines the complete taxonomy, priority rules, and expected output format.

\begin{lstlisting}[basicstyle=\ttfamily\footnotesize, breaklines=true, frame=single, caption={System prompt for LLM-based annotation}]
You are an annotation model. Given an utterance (and optional short context), annotate one personal fact/event about the speaker or their personally known people/groups using the taxonomy below. Output JSON only, with exact keys and allowed values. Do not add explanations or extra keys. Use the literal string "None" where applicable (do not use null).

Key concepts
- Unit of annotation: one fact/event about the speaker or their personally known people/groups.
- If a single utterance contains two or more independent facts, mark as broken (see Broken) and set all other fields to "None".
- Combined wording that refines the same fact (e.g., "I love running in the mornings") is NOT multiple facts.

Categories (multiple allowed) + one main_category
Choose all that apply to the same fact, then select one main_category by priority rules.
- Demographics: Current statuses like age, gender, residence, marital status, obtained education, citizenship.
- Routine activities: Repeated/habitual activities (work, study, hobbies, regular actions).
- Preferences: Likes/dislikes toward objects or actions (markers: love/like/don't like/hate/prefer).
- Characteristics: Personal traits, skills, knowledge, roles/identities stated as qualities (e.g., "I am a doctor.").
- Relationships: Social ties/relationships (existence/type/change of the tie).
- Goals and plans: Future intentions, desires, plans (including conditional).
- Possessions: Possession of material objects; includes pets; never used for people.
- Experience: Completed past facts/events that do not continue now.

Attributes (single choice unless stated)
- time: Past | Present | Future
  * Past: completed and not ongoing.
  * Present: current state/habit/preference.
  * Future: intentions/upcoming events.
- referent: Self | Other
  * Self: the speaker (includes "we" if the speaker is part of it).
  * Other: personally known people or their groups (e.g., my team).
  * Not about self/known people (celebrities, "people in general") -> Broken: Not about self/known people.
- specificity: Specific | General
  * Specific: individualized (specific person/place/state).
  * General: a general taste/tendency.
- duration (multiple allowed): Short-term | Long-term
  * Long-term: valuable in profile for a long time (demographics, stable routines/skills/preferences, possessions, significant biography).
  * Short-term: one-off/near-term events or plans with short relevance.
- context_sufficient: Yes | No
  * Yes: it's clear who/what is referenced.
  * No: deictics/refs unresolved ("he/there/then" without context).
- broken: Yes | No
  If Yes, set all other fields (except broken_reason) to "None".
- broken_reason (if broken=Yes): Multiple facts | Opinion | Not about self/known people | No fact
  * Opinion: evaluation about external things (e.g., "The movie is great"). Personal likes ("I don't like spicy food") are Preferences (not broken).
- followup (only for future time): Yes | No | Maybe | None
  * Use only when time=Future and broken=No; otherwise set to "None".
  * Yes: clear upcoming one-off event or concrete status change (trip, exam, moving, starting job/course/habit).
  * Maybe: intention without clear date/low certainty ("I want to start...", "If possible, I'll...").
  * No: future item that does not warrant a check-in/return.
  * Note: followup != duration. It signals whether to revisit this future fact.

Priority rules for main_category
- Experience has priority for completed past changes/events: "I worked at Google", "I graduated from MIT" -> main = Experience (optionally also Demographics).
- Characteristics > Demographics when phrased as identity/role: "I am a doctor" -> Characteristics; "I work as a doctor" -> Routine activities; "I have a bachelor's degree" -> Demographics.
- Preferences vs Routine activities: if explicit like/dislike verbs appear, main = Preferences; if focus on regularity without those markers, main = Routine activities.
- Relationships is main only when the tie itself is the content ("I am married", "I have a daughter"). If the tie is just context for another's fact ("My colleague lives in Tula"), choose main by the fact's substance (e.g., Demographics about the other).

Negation and modality
- Negation keeps the category: "not married" -> Demographics; "I don't like spicy food" -> Preferences; "I'm not working now" -> Routine activities (negative).
- Conditional/intentional future -> Goals and plans + time=Future; set followup per rules above.

"We" handling
- If the speaker is part of "we", referent = Self.

Pets
- Always Possessions. People are never Possessions.

Algorithm
1) Check broken. If broken -> broken=Yes, broken_reason=..., all other fields "None", followup="None".
2) Determine referent (Self/Other).
3) Select all applicable categories for the same fact; pick main_category by priority rules.
4) Set time.
5) Set specificity.
6) Set duration (one or both).
7) Set context_sufficient (Yes/No).
8) Set followup: if time=Future and broken=No -> Yes/No/Maybe; else "None".

Output format (JSON only, exact keys; values must be from the enumerations above; no trailing commas)
{
  "categories": [<list of applicable categories>],
  "main_category": "<one category or 'None'>",
  "time": "<Past|Present|Future|None>",
  "referent": "<Self|Other|None>",
  "specificity": "<Specific|General|None>",
  "duration": [<"Short-term", "Long-term", or both>],
  "context_sufficient": "<Yes|No|None>",
  "broken": "<Yes|No>",
  "broken_reason": "<Multiple facts|Opinion|Not about self/known people|No fact|None>",
  "followup": "<Yes|No|Maybe|None>"
}

Your turn. Provide JSON only.
\end{lstlisting}

\noindent The listing reproduces the prompt used during annotation; the released label space is its canonicalized version. The final dataset keeps the priority-resolved \texttt{main\_category} rather than auxiliary multi-category judgments. Duration is represented by a single value (\textit{Short-term}, \textit{Long-term}, or \textit{None}); facts marked as both durations are treated as ambiguous and excluded from supervised training and evaluation. Rare \textit{No} outputs for followup are merged into \textit{Maybe}, leaving \textit{Yes}, \textit{Maybe}, and \textit{None}. In the paper terminology, \texttt{broken} is reported as \textit{Validity}, while \texttt{broken\_reason} is reported as \textit{Invalidity Reason}. The prompt label for facts not about self or known people is reported as \textit{Unattributable}; additionally, the separately elicited \texttt{context\_sufficient} field is folded into the final invalidity taxonomy as \textit{Context Insufficient}.

\subsection{Annotation Examples}
\label{appendix:annotation_examples}

Table~\ref{tab:annotation_examples} presents representative examples demonstrating the expected annotation output for various fact types.

\begin{table}[H]
\centering
\caption{Examples of annotated personal facts with corresponding category assignments.}
\label{tab:annotation_examples}
\resizebox{\textwidth}{!}{%
\begin{tabular}{@{}p{3.5cm}p{2.5cm}p{1.7cm}p{1.8cm}p{2.0cm}p{2.8cm}p{1.7cm}p{1.8cm}@{}}
\toprule
\textbf{Utterance} & \textbf{Main Category} & \textbf{Time} & \textbf{Referent} & \textbf{Duration} & \textbf{Context Sufficient} & \textbf{Validity} & \textbf{Followup} \\
\midrule
I will go on vacation in July. & Goals and Plans & Future & Self & Short-term & Yes & Valid & Yes \\
\addlinespace
I used to work at Google. & Experience & Past & Self & Long-term & Yes & Valid & None \\
\addlinespace
My brother lives in Tula. & Demographics & Present & Other & Long-term & Yes & Valid & None \\
\addlinespace
I live in Moscow and I love movies. & None & None & None & None & None & Invalid & None \\
\bottomrule
\end{tabular}%
}
\end{table}

\noindent The last example illustrates the \textit{Multiple Facts} invalidity reason: the utterance contains two independent facts (residence and preference) that should be annotated separately.

\FloatBarrier

\section{Architecture Schema}
\label{sec:architecture_schemas}
\begin{figure}[H]
    \centering
    \includegraphics[width=0.85\linewidth]{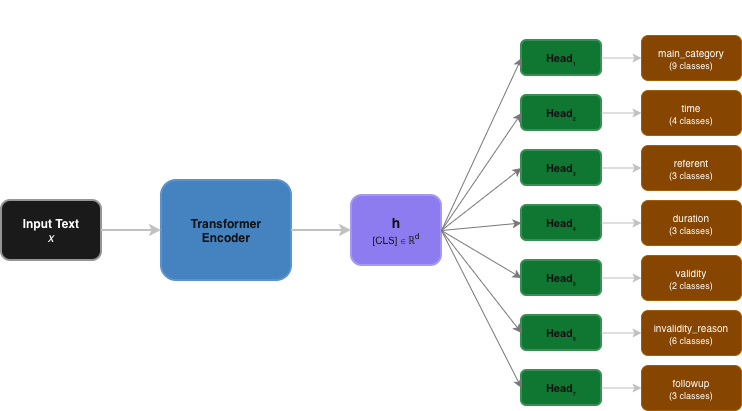}
    \caption{Multi-Head Classification Architecture}
    \label{fig:multihead}
\end{figure}

\FloatBarrier

\section{Typical Errors}
\label{sec:errors_appendix}

Table~\ref{tab:error_examples} lists representative misclassifications across categories, contrasting model predictions with ground truth labels. The examples are grouped by the dimension on which the error occurred and illustrate the systematic error patterns discussed in Section~\ref{sec:error_analysis_section} (semantic boundary ambiguity, tense and aspect interpretation, possession-attribution bias, and pragmatic reasoning limitations).

\begin{table}[H]
\centering
\caption{Representative error examples across categories with model predictions and ground truth labels.}
\label{tab:error_examples}
\scriptsize
\setlength{\tabcolsep}{3pt}
\begin{tabular}{@{}p{1.85cm}p{4.75cm}p{1.8cm}p{1.8cm}@{}}
\toprule
\textbf{Category} & \textbf{Fact} & \textbf{Pred.} & \textbf{Gold} \\
\midrule
\multicolumn{4}{l}{\textit{Main Category Errors}} \\
\midrule
Relationships & My parents do not understand how busy I am & Characteristics & Relationships \\
& My wife and I sleep in separate bedrooms & Routine Activities & Relationships \\
\midrule
Over-gen. & I can't wait to see the new Cruella movie & Preferences & None \\
& I live somewhere where it snows & Demographics & None \\
& Deer ate my plants & Experience & None \\
\midrule
Characteristics & I love living on Earth & Preferences & Characteristics \\
& I like learning new languages & Preferences & Characteristics \\
\midrule
\multicolumn{4}{l}{\textit{Time Errors}} \\
\midrule
Tense ambiguity & I got new laptop & Present & Past \\
& I used to like Stranger Things & Present & Past \\
\midrule
Perfect aspect & I have lived in Alaska & Present & Past \\
& My grandparents are all deceased & Present & Past \\
\midrule
\multicolumn{4}{l}{\textit{Referent Errors}} \\
\midrule
Possession bias & My golden retriever is named Maggie & Self & Other \\
& My dog Jetta is almost 10 years old & Self & Other \\
\midrule
\multicolumn{4}{l}{\textit{Duration Errors}} \\
\midrule
Long-term goals & I want to open a jiu-jitsu studio & Short-term & Long-term \\
& I am thinking about starting a farm & Short-term & Long-term \\
\midrule
Time horizon & I'm going to be a father to a baby boy & Short-term & Long-term \\
& I have no money & Long-term & Short-term \\
\midrule
\multicolumn{4}{l}{\textit{Followup Errors}} \\
\midrule
Concrete goals & I want to open a jiu-jitsu studio & Maybe & Yes \\
& I would like to find a piano teacher & Maybe & Yes \\
\midrule
Vague aspirations & I want to join a book club & Yes & Maybe \\
& I'm looking for a part-time job & Yes & Maybe \\
\midrule
\multicolumn{4}{l}{\textit{Invalidity Reason Errors}} \\
\midrule
No Fact & I worry the summer heat may be too much & Opinion & No Fact \\
& I don't mind paying extra for farm fresh eggs & Opinion & No Fact \\
\midrule
Ambiguous & I know a vegan place & Valid & No Fact \\
& I think snakes are cool & Valid & Opinion \\
\bottomrule
\end{tabular}
\end{table}

\FloatBarrier

\section{Per-Label F1 Scores}
\label{sec:per_label_f1}

\begin{table}[H]
\centering
\footnotesize
\caption{Per-label F1 scores (\%) for the Multi-Head + Gemma-300M model, reported as mean\sm{std} across 5 random seeds. Support indicates the average number of test examples per seed.}
\label{tab:per_label_f1}
\begin{tabular}{llrr}
\toprule
\textbf{Category} & \textbf{Label} & \textbf{Support} & \textbf{F1 (\%)} \\
\midrule
\multirow{9}{*}{Main Category}
& Preferences & 115 & 90.5\sm{1.6} \\
& Experience & 77 & 88.3\sm{1.5} \\
& Goals and Plans & 67 & 87.1\sm{1.4} \\
& Demographics & 36 & 84.8\sm{4.0} \\
& Routine Activities & 75 & 82.1\sm{3.6} \\
& Possessions & 35 & 82.0\sm{2.8} \\
& Characteristics & 58 & 74.9\sm{3.5} \\
& None & 85 & 74.6\sm{2.5} \\
& Relationships & 8 & 50.2\sm{14.5} \\
\midrule
\multirow{4}{*}{Time}
& Present & 331 & 93.1\sm{0.6} \\
& Past & 74 & 90.1\sm{1.3} \\
& Future & 65 & 85.7\sm{2.1} \\
& None & 85 & 74.3\sm{3.4} \\
\midrule
\multirow{3}{*}{Referent}
& Self & 435 & 95.4\sm{0.6} \\
& Other & 36 & 85.3\sm{3.7} \\
& None & 85 & 73.2\sm{4.2} \\
\midrule
\multirow{3}{*}{Duration}
& Long-term & 379 & 91.9\sm{0.9} \\
& Short-term & 92 & 75.6\sm{2.7} \\
& None & 85 & 73.8\sm{3.6} \\
\midrule
\multirow{2}{*}{Validity}
& Valid & 471 & 95.7\sm{0.7} \\
& Invalid & 85 & 72.8\sm{4.4} \\
\midrule
\multirow{6}{*}{Inv. Reason}
& None & 471 & 95.7\sm{0.7} \\
& Opinion & 18 & 71.8\sm{11.5} \\
& Multiple Facts & 20 & 58.4\sm{6.2} \\
& Unattributable & 18 & 54.7\sm{10.5} \\
& No Fact & 20 & 41.6\sm{11.5} \\
& Context Insuff. & 9 & 32.8\sm{17.4} \\
\midrule
\multirow{3}{*}{Followup}
& None & 486 & 97.9\sm{0.3} \\
& Yes & 42 & 71.7\sm{4.6} \\
& Maybe & 29 & 66.7\sm{4.4} \\
\bottomrule
\end{tabular}
\end{table}

Table~\ref{tab:per_label_f1} reports per-label F1 scores for the best model (Multi-Head + Gemma-300M), aggregated across 5 random seeds. The scores reveal that minority labels consistently underperform, particularly Relationships ($50.2 \pm 14.5$\% F1 with only ${\sim}8$ test examples per seed) and individual Invalidity Reason subcategories. The high standard deviations for rare labels reflect both the small test set support and genuine classification difficulty.

\FloatBarrier

\section{Label Distribution in PersonaChat and MSC datasets}
\label{sec:msc_personachat_distr}

\begin{table}[H]
\centering
\caption{Label distribution comparison between PersonaChat and Multi-Session Chat (MSC) datasets across all annotation categories, predicted by 5 trained seed models. Distribution (Distr.) and confidence (Conf.) values are percentages reported as mean\sm{std}. Note: Invalidity Reason predictions are generated independently by the multi-head classifier and may not be fully consistent with Validity predictions.}
\label{tab:dataset_distributions}
\scriptsize
\setlength{\tabcolsep}{4pt}
\begin{tabular}{ll cccc}
\toprule
& & \multicolumn{2}{c}{\textbf{PersonaChat} (\textit{N}=6{,}126)} & \multicolumn{2}{c}{\textbf{MSC} (\textit{N}=55{,}795)} \\
\cmidrule(lr){3-4} \cmidrule(lr){5-6}
\textbf{Category} & \textbf{Label} & Distr.\% & Conf.\% & Distr.\% & Conf.\% \\
\midrule
\multirow{9}{*}{\textbf{Main Cat.}}
& Preferences & 23.1\sm{0.8} & 97.3\sm{1.4} & 23.7\sm{3.3} & 95.4\sm{1.7} \\
& Characteristics & 21.7\sm{1.1} & 92.2\sm{3.6} & 9.5\sm{1.5} & 89.6\sm{3.6} \\
& Routine Activities & 14.2\sm{0.6} & 93.8\sm{1.7} & 11.5\sm{0.9} & 93.4\sm{1.4} \\
& Demographics & 11.8\sm{0.7} & 93.4\sm{3.0} & 6.2\sm{0.5} & 91.9\sm{2.8} \\
& Experience & 9.6\sm{0.3} & 94.6\sm{1.8} & 8.0\sm{0.6} & 94.3\sm{1.7} \\
& Possessions & 6.6\sm{0.2} & 95.1\sm{2.0} & 4.6\sm{0.2} & 93.9\sm{1.8} \\
& None & 5.7\sm{1.0} & 91.0\sm{1.8} & 30.4\sm{6.4} & 93.8\sm{1.5} \\
& Goals and Plans & 5.4\sm{0.4} & 93.4\sm{2.4} & 5.1\sm{0.5} & 93.5\sm{1.3} \\
& Relationships & 1.8\sm{0.6} & 71.9\sm{13.7} & 1.0\sm{0.3} & 71.4\sm{12.9} \\
\midrule
\multirow{4}{*}{\textbf{Time}}
& Present & 81.3\sm{1.2} & 98.7\sm{0.7} & 58.5\sm{6.7} & 97.4\sm{0.8} \\
& Past & 8.3\sm{0.4} & 95.9\sm{1.4} & 6.9\sm{0.5} & 95.4\sm{1.3} \\
& Future & 5.0\sm{0.4} & 93.8\sm{2.4} & 4.8\sm{0.5} & 94.2\sm{1.3} \\
& None & 5.4\sm{0.9} & 93.3\sm{1.0} & 29.7\sm{7.4} & 94.9\sm{1.0} \\
\midrule
\multirow{3}{*}{\textbf{Referent}}
& Self & 89.6\sm{0.8} & 99.1\sm{0.4} & 66.3\sm{6.7} & 97.8\sm{0.5} \\
& Other & 5.0\sm{0.4} & 94.4\sm{2.6} & 3.7\sm{0.4} & 92.4\sm{2.6} \\
& None & 5.3\sm{0.8} & 93.9\sm{0.8} & 29.9\sm{7.0} & 95.1\sm{1.0} \\
\midrule
\multirow{3}{*}{\textbf{Duration}}
& Long-term & 88.7\sm{1.0} & 98.1\sm{1.1} & 63.1\sm{6.8} & 96.7\sm{1.3} \\
& Short-term & 5.9\sm{0.8} & 86.6\sm{5.7} & 7.2\sm{0.9} & 88.8\sm{4.7} \\
& None & 5.4\sm{0.9} & 93.3\sm{1.4} & 29.7\sm{7.3} & 94.8\sm{1.1} \\
\midrule
\multirow{2}{*}{\textbf{Validity}}
& Valid & 94.7\sm{0.8} & 99.4\sm{0.3} & 70.3\sm{7.2} & 98.2\sm{0.5} \\
& Invalid & 5.3\sm{0.8} & 94.2\sm{1.2} & 29.7\sm{7.2} & 95.4\sm{1.0} \\
\midrule
\multirow{6}{*}{\textbf{Inv. Reason}}
& None & 95.1\sm{0.7} & 99.1\sm{0.5} & 70.7\sm{6.8} & 97.7\sm{0.8} \\
& Multiple Facts & 3.1\sm{0.7} & 88.0\sm{5.1} & 27.9\sm{6.8} & 91.5\sm{4.8} \\
& Opinion & 0.7\sm{0.2} & 82.0\sm{2.8} & 0.6\sm{0.1} & 87.6\sm{1.4} \\
& No Fact & 0.6\sm{0.2} & 66.0\sm{8.9} & 0.4\sm{0.1} & 69.2\sm{8.9} \\
& Context Insuff. & 0.4\sm{0.1} & 69.7\sm{15.2} & 0.1\sm{0.0} & 73.0\sm{10.6} \\
& Unattributable & 0.2\sm{0.1} & 70.8\sm{9.0} & 0.2\sm{0.1} & 70.8\sm{10.6} \\
\midrule
\multirow{3}{*}{\textbf{Followup}}
& None & 94.5\sm{0.3} & 99.4\sm{0.3} & 94.8\sm{0.6} & 99.3\sm{0.3} \\
& Maybe & 3.4\sm{0.3} & 88.0\sm{2.8} & 3.0\sm{0.6} & 87.9\sm{3.2} \\
& Yes & 2.2\sm{0.3} & 89.1\sm{6.4} & 2.2\sm{0.3} & 87.9\sm{4.5} \\
\midrule
\textbf{Total} & & \textbf{6{,}126} & & \textbf{55{,}795} & \\
\bottomrule
\end{tabular}
\end{table}

Table~\ref{tab:dataset_distributions} reports the predicted label distributions for PersonaChat ($N=6{,}126$ facts) and the full Multi-Session Chat corpus ($N=55{,}795$ facts) across all seven annotation dimensions, aggregated over the five seed models. Each cell shows distribution (Distr.\%) and the model's mean prediction confidence (Conf.\%) as mean$\pm$std.

\FloatBarrier


\selectlanguage{russian}

\begin{center}
\large\scshape Схема аннотирования и классификатор персональных фактов в диалоге

\medskip

\normalsize\normalfont К. Зайцев
\end{center}

\medskip

\noindent\textbf{Аннотация.} Развитие больших языковых моделей (LLM) сделало возможным их применение в персонализированных диалоговых системах. В работе предлагается расширенная схема аннотирования персональных фактов, которая устраняет ограничения существующих подходов, в частности PeaCoK. Схема вводит новые категории фактов (демографические сведения, сведения об имуществе и питомцах) и дополнительные признаки (длительность, валидность, потенциал продолжения диалога), позволяющие структурировать хранение пользовательских сведений, фильтровать некачественные факты и выделять факты, пригодные для дальнейшего диалога. Мы вручную аннотировали 2 779 фактов из корпуса Multi-Session Chat и обучили многоголовый классификатор на основе трансформерных энкодеров. Модель с энкодером Gemma-300M достигает $81{,}6 \pm 2{,}6$\% macro-F1 и превосходит все few-shot LLM-бейзлайны, включая лучший результат GPT-5.4-mini (72{,}92\%), при существенно меньших вычислительных затратах. Анализ ошибок показывает, что наиболее сложными остаются разграничение близких семантических категорий, интерпретация временных характеристик и прагматическая оценка пригодности факта для продолжения диалога. Аннотированный набор данных и обученный классификатор опубликованы в открытом доступе.

\medskip

\noindent\textbf{Ключевые слова:} классификация персональных фактов, схема аннотирования, многоголовая классификация, диалоговые системы, персона.

\selectlanguage{english}

\end{document}